# 3DFR: A Swift 3D Feature Reductionist Framework for Scene Independent Change Detection

Murari Mandal, Vansh Dhar, Abhishek Mishra, Santosh Kumar Vipparthi

*Abstract*—In this paper we propose an end-to-end swift 3D feature reductionist framework (3DFR) for scene independent change detection. The 3DFR framework consists of three feature streams: a swift 3D feature reductionist stream (AvFeat), a contemporary feature stream (ConFeat) and a temporal median feature map. These multilateral foreground/background features are further refined through an encoder-decoder network. As a result, the proposed framework not only detects temporal changes but also learns high-level appearance features. Thus, it incorporates the object semantics for effective change detection. Furthermore, the proposed framework is validated through a scene independent evaluation scheme in order to demonstrate the robustness and generalization capability of the network. The performance of the proposed method is evaluated on the benchmark CDnet 2014 dataset. The experimental results show that the proposed 3DFR network outperforms the state-of-the-art approaches.

## I. INTRODUCTION

CHANGE detection in videos is one of the elementary visual analytics tasks. The applications of change detection pervade patient monitoring, depression analysis, dynamic gesture recognition, gait analysis, moving object detection, anomaly detection, object tracking, action recognition, behavior analysis, etc. Change detection techniques in the literature can be loosely categorized into parametric and non-parametric approaches. Stauffer and Grimson [1] developed parametric Gaussian Mixture Models (GMM) which models the statistical distribution of intensities at each pixel location. Many variations of this approach [8, 9] have been presented in the literature. However, most of the modern works [2-5] in change detection are inspired by the nonparametric background modeling approaches given in [6] and [7]. Wang and Suter [6] presented a consensus-based method to collect recent history frames as background samples. These samples were updated through a first-in-first-out policy. Furthermore, Barnich and Droogenbroeck [7] proposed three important strategies for background subtraction: random sample replacement, memoryless update policy, and spatial diffusion via neighborhood background sample propagation. Hofmann et al. [2] devised a mechanism to adaptively update the pixel-wise decision thresholds and update rate. St-Charles et al. [4, 5] proposed a more sophisticated algorithm SuBSENSE using spatiotemporal feature descriptors LBSP and adaptive feedback mechanism. Various adaptations of SuBSENSE [11, 12] have also been proposed to further improve the performance. A deterministic background model update policy was proposed by Mandal et al. [10]. Multiple background models based on the fusion of RGB and YCbCr color models [13], background word consensus [14] and semantic background subtraction [15] are some other interesting methods proposed by the researchers in recent times. Bianco et al. [5] conducted multiple experiments to combine various change detection techniques through genetic programming for improved performance.

In recent years, advances in deep learning have led to the development of various convolutional neural networks (CNNs) based techniques for change detection. Many attempts in this domain leverage off-the-shelf pre-trained CNNs and integrate it with designed background modeling techniques for temporal feature encoding [16-19]. Certain researchers [16, 20-23] have divided the frames and background model/ recent history frames into patches. Further, they trained the CNN models with concatenated patches as input. Chen at al. [24] designed an attention ConvLSTM to model pixel-wise changes over time. Patil and Murala [25] designed a compact encoder-decoder to extract the foreground regions by modelling saliency using a small video stream. Furthermore, Yang et al. [26] used the fully convolutional network (FCN) with skip connections and atrous convolutions to extract spatial information. Bakkay et al. [27] designed a conditional Generative Adversarial Network (cGAN) to learn the motion features for background subtraction.

Most of the existing CNN based methods in the literature are dependent on the hand-crafted approaches to estimate the background. Moreover, the features are primarily encoded using 2D CNN. Although, these techniques work very well over certain scenarios, however, in order to extract generalizable motion features for a variety of scenes, it is important to encode spatiotemporal features from video streams.

Moreover, the abovementioned supervised learning methods [16-27] for change detection have been evaluated without ensuring scene independence in the train-test data division.

Murari Mandal and Santosh Kumar Vipparthi are with Vision Intelligence Lab, Department of Computer Science and Engineering, MNIT Jaipur, India (Email: murarimandal.cv@gmail.com; skvipparthi@mnit.ac.in). The work was supported by the project #SERB/F/9507/2017.

Vansh Dhar and Abhishek Mishra are with School of Computing and Information Technology, Manipal University, Jaipur (Email: vanshdhar.ai@gmail.com; mishra.abhishek.ai@gmail.com)

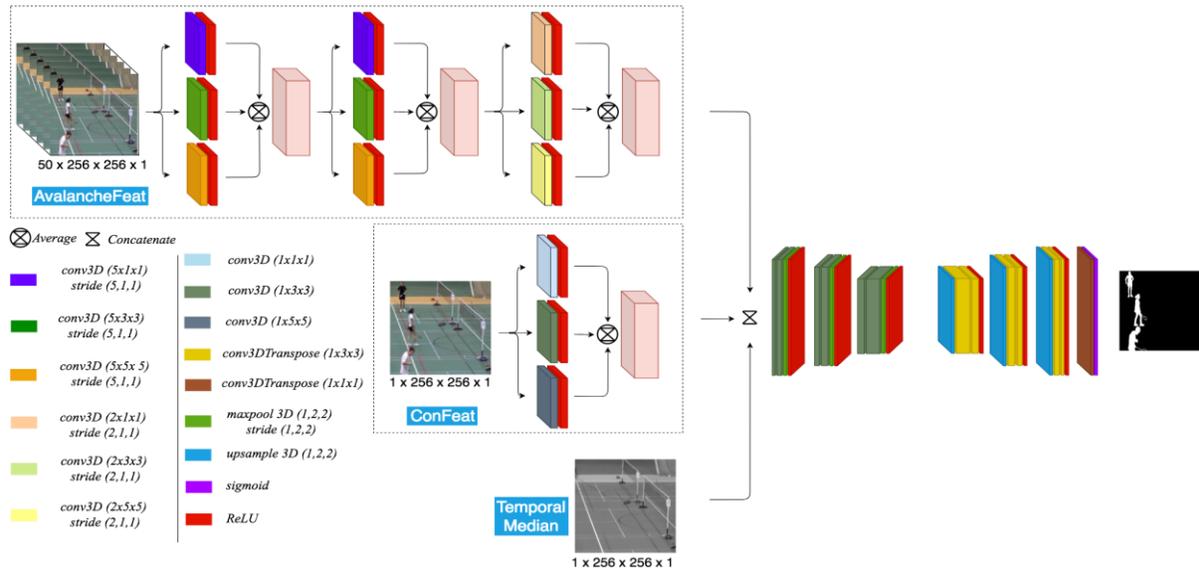

Fig. 1. The proposed 3DFR framework. The network takes three inputs: recent history frames, current frame and temporal median of the history frames. In our experiments, we used 50 recent history frames for training and evaluation.

By scene independence, we mean that the train set should not contain partial data of the videos used in the test set. The intuition behind this is that the background remains more or less similar in the entire video sequences. Therefore, even though the train and test set (of the same video) might contain different frames but they almost have the same background. This would give an unfair advantage to the CNN model in comparison to unsupervised models and lead to inconsistent comparative analysis. This makes it extremely difficult to assess the generalization capabilities of the corresponding methods.

In this paper, we propose an end-to-end 3D CNN based swift spatiotemporal feature reductionist framework 3DFR for scene independent change detection. The proposed framework enables background subtraction using spatiotemporal feature reductionist encoding, contemporary feature extraction, and temporal median. The change detection is performed by designing a multi-lateral feature-based encoder-decoder scheme. The complete framework is trained in an end-to-end manner without any data preprocessing. To summarize, this paper makes the following contributions:
1. In 3DFR framework, we introduced three lateral feature encoding schemes named AvFeat, ConFeat and temporal median.
2. The AvFeat block is designed to learn the spatiotemporal features through a swift feature reductionist approach. By 'swift', we refer to a rapid reduction in temporal depth in order to extract the spatiotemporal variations in recent history. In order to preserve the salient motion information, we used features from multiple receptive fields at each layer of the AvFeat block.
3. In addition, the ConFeat block represents the salient feature encoded from the current frame by extracting features from multiple receptive fields. Moreover, the temporal median from recent observations is extracted to offer attention to relevant background regions.
4. The salient foreground/background features from three different streams are refined through an encoder-decoder network to generate a robust motion segmentation map.

## II. PROPOSED 3DFR FRAMEWORK

The proposed 3DFR framework is based on four modules as shown in Fig. 1. In the following subsections, we give a detailed description of these different constituent modules. We also present an analysis of 3DFR network for its effectiveness in unseen videos.

### A. Avalanche Feature (AvFeat) Block

Inspired by the avalanche effect in cryptography, we design the AvFeat block for motion feature extraction using 3D convolutional kernels. Avalanche effect in cryptography [28] refers to the property, that if an input is changed slightly, the output changes significantly. In terms of video frames, the objective is to capture motion information efficiently at each pixel location corresponding to the raw pixel-level changes over time.

We rapidly (in three stages) decrease the temporal depth by adopting a swift spatiotemporal feature reductionist approach. In the initial two reduction stages, features from multiple receptive fields are encoded by averaging the feature responses from receptive fields of size 5x1x1, 5x3x3 and 5x5x5. Similarly, receptive fields of size 2x1x1, 2x3x3 and 2x5x5 are used in the final reduction stage. The temporal depth reduction is performed by the factor of 5, 5 and 2 in three stages respectively. This ensures robust background estimation through feature learning in both local and global context by progressive elimination of temporal movements as shown in Fig. 1. Moreover, the resultant response represents the features estimated from multiple receptive contexts in both spatial and temporal dimensions. For an input tensor $T_M$, we compute AvFeat features $AV_M$ using Eq. (1).

$$AV_M = \zeta_3(\zeta_2(\zeta_1(T_M)))  \quad (1)$$

where $\zeta_1(\cdot)$, $\zeta_2(\cdot)$ and $\zeta_3(\cdot)$ compute the averaging features from multiple receptive fields with each using 8 kernels and strides, (5, 1, 1), (5, 1, 1) and (2, 1, 1), respectively. From $M$ historical frames, the AvFeat block generates eight motion entropy feature maps.

## B. ConFeat Block

In order to delineate semantically accurate shape representation for change detection, we extract contemporary features (ConFeat) from the current frame. This ensures a more accurate shape representation of the foreground regions. The ConFeat for the current frame is computed by averaging the feature responses from receptive fields of size 1x1x1, 1x3x3 and 1x5x5 with each using 8 kernels.

## C. Temporal Median

Moreover, to represent the coarse background patterns, we also extract the pixel-wise temporal median of $M$ recent frames. This enhances the robustness of background features estimated from the AvFeat block. The temporal median $TM$ is calculated using in Eq. (2).

$$TM(x) = median(I_k(x)|_{k=1}^M) \quad (2)$$

where $x$ represents the pixel location in the image. Finally, these multi-stream features are concatenated using Eq. (3).

$$MSFeat = [AV_M, ConFeat, TM] \quad (3)$$

## D. Multi-lateral feature-based Encoder-Decoder

The salient foreground/background features generated by $ConFeat$, $TM$ and $AV_M$ are refined through an encoder-decoder network to generate the final segmentation map. The proposed encoder (*MLEn*)-decoder (*MLDec*) assists in learning high-level appearance features by incorporating object semantics for effective change detection. The *MLEn* and *MLDec* are defined through the following equations.

$$MLEn = Enc_3(Enc_2(Enc_1(MSFeat))) \quad (4)$$

$$Enc_k(z) = \Re(mp_{1,2,2}(\kappa_{2^{k+3},1,h,w} \otimes (\kappa_{2^{k+2},1,h,w} \otimes z))) \quad (5)$$

$$MLDec = Dec_3(Dec_2(Dec_1(MLEn))) \quad (6)$$

$$Dec_k(z) = \Re(\kappa_{2^{6-k},1,h,w}^T \otimes (\kappa_{2^{7-k},1,h,w}^T \otimes up_{1,2,2}(z))) \quad (7)$$

where $mp_{1,2,2}$, $up_{1,2,2}$, $\kappa_{x,1,h,w}^T$ denote max pooling, up-sampling, and transposed convolutional kernel respectively. The max pool and up sample operations are applied with stride=2. For all the convolutional and transpose convolutional operations, we used h=3, w=3, and stride=1. The change probability map is predicted as given in Eq. (8).

$$FG = \delta(\kappa_{1,1,1,1}^T \otimes MLDec) \quad (8)$$

where $\delta(\cdot)$ denote the sigmoid function.

## E. Analysis of 3DFR Network

In terms of background estimation (BE), the CNN based approaches in [16, 19-22] are dependent on handcrafted algorithms to extract the temporal features. Whereas, the authors in [17, 18] have just performed frame-level segmentation without considering the historical context. However, in the presented 3DFR, we proposed AvFeat block to model temporal features from the recent history for effective BE. Moreover, to highlight the dissimilarity between foreground-background and characterize the change information, the contrasting features between the estimated backgrounds (AvFeat and temporal median) are further concatenated with the current frame features (ConFeat). This intuitive approach helps the model to learn the underlying problem of change detection. Therefore, our model performs well even on unseen videos rather than overfit to perform well only on the same videos used in training. Moreover, we would like to highlight the fact that not many existing papers have evaluated their deep learning models on scene independent setup. Which makes it is very difficult to evaluate their robustness in unseen videos. The proposed 3DFR also offers a customizable framework to design and develop 3D CNN based techniques for change detection.

TABLE I
COMPARATIVE ANALYSIS OF EVALUATION SCHEMES ADOPTED BY EXISTING DEEP LEARNING APPROACHES AND THE PROPOSED 3DFR FRAMEWORK.

| Method | Train-Test Division | SIE |
|---|---|---|
| Chen et al. [24] | Temporal Division | No |
| Yang et al. [26] | Temporal Division | No |
| Babaee et al. [21] | Temporal Division | No |
| Nguyen et al. [20] | Random Division | No |
| Lin et al. [16] | Leave-one-video-out | Yes |
| Lim et al. [17] | Selective Division | No |
| Zeng et al. [18] | Random Division | No |
| Lim et al. [19] | Leave-one-video-out | No |
| Bakkay et al. [27] | Temporal Division | No |
| Brahman et al. [22] | Temporal Division | No |
| Patil and Murala [25] | Random Division | No |
| **Proposed 3DFR** | **Leave-one-video-out** | **Yes** |

*SIE.: Scene independent evaluation.

## III. EXPERIMENTAL SETUP, RESULTS & DISCUSSIONS

We evaluated our method on the benchmark CDnet 2014 [29] dataset. The CDnet dataset contains 53 annotated videos from numerous real-world scenarios. These videos are categorized into 11 categories. The performance is measured in terms of F-score. We have compared our work with nine recent state-of-the-art change detection methods.

## A. Need for Scene Independent Evaluation in Deep Learning

The CNN based methods for change detection in the literature [16-27] have evaluated their models in various train-test divisions without properly ensuring scene independence. For example, the most commonly adopted scheme is to divide the frame sequences from a single video into training and testing

TABLE II
SCENE INDEPENDENT DATA DIVISION USED IN 3DFR EXPERIMENTS

| Category | Train Data [no. of frames] | Test Data [no. of frames] |
|---|---|---|
| Bad Weather | skating, snowfall, wetsnow [5,900] | blizzard [3,050] |
| Baseline | highway, office, PETS2006 [3,613] | pedestrians [800] |
| Camera Jitter | badminton, boulevard, sidewalk [2,316] | traffic [622] |
| Dynamic Background | canoe, fall, fountain01, fountain02, overpass [7,177] | boats [6100] |
| Intermittent Object Motion | abandonedBox, sofa, streetLight, tramstop, winterDriveway [10,710] | parking [1401] |
| Low Frame Rate | port_0_17fps, tramCrossroad_1fps, tunelExit_0_35fps [2,250] | turnpike_0_5fps [350] |
| Night Videos | bridgeEntry, busyBoulvard, fluidHighway, streetCornerAtNight, winterStreet [4642] | tramStation [1201] |
| Shadow | backdoor, bungalows, cubicle, peopleInShade [10057] | busStation [951] |
| Thermal | diningRoom, lakeSide, library, park [13,154] | corridor [4901] |
| Turbulence | turbulence1, turbulence2, turbulence3 [4,700] | tubulence1 [1,400] |

TABLE III
CHANGE DETECTION PERFORMANCE COMPARISON OF THE PROPOSED AND EXISTING STATE-OF-THE-ART APPROACHES ON CDNET 2014 DATASET IN SCENE INDEPENDENT EVALUATION SETUP. (THE BEST F-SCORES ARE HIGHLIGHTED IN BOLD)

| Method | S.I.E. | Bliz | Ped | Traffic | Boat | Parking | Turp | TrS | BuS | Cor | Tu1 | Overall |
|---|---|---|---|---|---|---|---|---|---|---|---|---|
| SuBSense [4] | Yes | 0.85 | 0.95 | 0.80 | 0.69 | 0.48 | 0.85 | 0.76 | 0.86 | 0.91 | 0.79 | 0.80 |
| VIBE [7] | Yes | 0.53 | 0.90 | 0.66 | 0.22 | 0.26 | 0.60 | 0.62 | 0.67 | 0.75 | 0.58 | 0.58 |
| PAWCS [14] | Yes | 0.66 | 0.95 | 0.83 | 0.88 | 0.21 | 0.91 | 0.63 | 0.86 | 0.65 | 0.68 | 0.73 |
| IUTIS-5 [5] | Yes | 0.80 | 0.97 | 0.83 | 0.75 | 0.65 | 0.89 | 0.75 | 0.87 | 0.90 | 0.63 | 0.80 |
| UBSS [13] | Yes | 0.86 | 0.96 | 0.68 | 0.90 | 0.62 | 0.89 | 0.74 | 0.87 | **0.92** | 0.54 | 0.80 |
| WeSamBe [11] | Yes | 0.86 | 0.96 | 0.80 | 0.64 | 0.41 | 0.91 | 0.79 | 0.86 | 0.89 | 0.71 | 0.78 |
| SemanticBGS [15] | Yes | 0.84 | **0.98** | 0.84 | **0.98** | 0.69 | 0.88 | 0.71 | 0.92 | 0.82 | 0.30 | 0.80 |
| DeepBS$ [21] | No | 0.61 | 0.95 | **0.88** | 0.81 | 0.60 | 0.49 | 0.16 | **0.94** | 0.89 | 0.77 | 0.71 |
| MSFgNet$ [25] | No | 0.85 | 0.94 | 0.85 | 0.80 | **0.87** | 0.91 | **0.86** | 0.89 | 0.80 | 0.76 | 0.85 |
| **3DFR** | Yes | **0.95** | 0.93 | 0.65 | 0.90 | **0.87** | **0.92** | 0.79 | 0.79 | 0.89 | **0.86** | **0.86** |

*S.I.E.: Scene Independent Evaluation, Bliz: Blizzard (Bad Weather), Ped: Pedestrian (Baseline), Traffic (Camera Jitter), Boat (Dynamic Background), Parking (Intermittent Object Motion), Turp: Turnpike_0_5fps (Low Frame Rate), TrS: Tram station (Night Videos), BuS: Bus station (Shadow), Cor: Corridor (Thermal), Tu1: Turbulence01(Turbulence). $These results are collected from the original paper and do not adhere to scene independent evaluation setup.

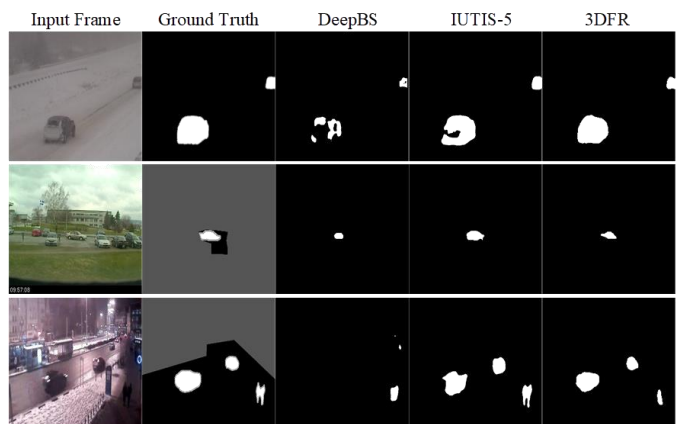

Fig. 2. Qualitative results of the proposed 3DFR and existing state-of-the-art approaches.

group with certain temporal proportions (50%-50% [22, 24, 27], 90%-10% [26]). Moreover, few techniques [17, 18, 20, 21] have either manually or randomly selected certain frames from each video to train the network and then reported the results for the same videos. The detailed description of the evaluation schemes adopted in the literature for deep learning-based change detection is given in Table I. Since, such schemes would clearly lead to very high similarity or dependency between the background scenes in train and test set. Thus, the existing approaches [17, 27] have achieved a very high F-Score in CDnet dataset. However, such evaluation schemes make it extremely difficult to assess the robustness and generalization of these deep learning approaches. To address the abovementioned problems, we propose a complete scene independent evaluation (SIE) scheme for change detection. In this scheme, we assign one video of a scene category as test data and use remaining videos as the training set. This ensures complete scene independence between the train and test sets. We give a detailed description of the train and test data division for our SIE setup in Table II. To the best of our knowledge, no baselines results are available for scene independent evaluation. Therefore, we have randomly selected one video from each category for evaluation. We exclude PTZ category due to camera motion. It consists of videos captured from moving cameras, two-position cameras, and variable zoom while recording. These causal factors are out of the scope of the proposed method. Also, in 'shadow' category, we did not include copyMachine video in the train set due to its diverging effect on the training process.

**Training Configuration.** The 3DFR framework is implemented using Keras library with TensorFlow backend and training is performed with batch size=1 over NVIDIA Titan Xp GPU system. We use binary cross-entropy loss function and SGD optimizer while training the network. The initial learning rate is set to 0.0006 which is further decreased by 0.0002 after every 20 epochs. The minimum learning rate is set to 0.0001.

*B. Results and Discussions*

The quantitative performance of the proposed and existing state-of-the-art techniques are given in Table III. The proposed network consists of 126.45K trainable parameters with a model size of 2.82 MB. The inference speed is 30 ms per frame or approximately 33.3 frames per second (FPS). The proposed 3DFR achieves an average F-score of 0.86 which is 6% higher than the next best hand-crafted approach. The proposed work outperforms DeepBS and MSFgNet by 15% and 1% respectively. Although, our proposed work was evaluated over scene independent setup, it still outperforms scene dependent results of DeepBS and MSFgNet. Thus, the proposed 3DFR framework is more robust and better generalized to handle unseen scenarios. We suggest that in order to demonstrate the generalization capability of supervised learning techniques, the evaluations must be done in such a scene independent setup. We also present a comparative qualitative analysis in Fig. 2. The results also show the potential of 3D convolution over 2D convolution [16-22, 25] in designing deep networks for change detection in videos.

IV. CONCLUSION

A novel 3D CNN based end-to-end deep learning framework is proposed for change detection in scene independent setup. We introduced AvFeat block using features from multiple receptive fields for swift spatiotemporal feature reduction. Moreover, a ConFeat block was proposed to delineate semantically accurate shape representations and the temporal median was extracted to represent the coarse background patterns. Change detection is performed by refining the salient foreground/background features through an encoder-decoder network. Furthermore, we present a scene independent train-test division strategy to promote the development of a more generalized feature learning framework for change detection. The proposed approach also outperforms existing state-of-the-art change detection techniques.

## REFERENCES

[1] C. Stauffer and W. E. L. Grimson, "Adaptive background mixture models for real-time tracking," in *Proc. IEEE Comput. Vis. Pattern Recognit.*, vol. 2, pp. 246-252, 1999.

[2] M. Hofmann, P. Tiefenbacher and G. Rigoll, "Background segmentation with feedback: The pixel-based adaptive segmenter," in *Proc. IEEE Comput. Vis. Pattern Recognit. Workshops*, pp. 38-43, 2012.

[3] P. L. St-Charles and G. A. Bilodeau, "Improving background subtraction using local binary similarity patterns," in *Proc. IEEE Winter Conf. Appl. Comput. Vis.*, pp. 509-515, 2014.

[4] P. L. St-Charles, G. A. Bilodeau and R. Bergevin, "SuBSENSE: A universal change detection method with local adaptive sensitivity," *IEEE Trans. Image Process.*, vol. 24, no. 1, pp. 359-373, 2015.

[5] S. Bianco, G. Ciocca and R. Schettini, "Combination of video change detection algorithms by genetic programming," *IEEE Trans. Evolutionary Computation*, vol. 21, no. 6, pp. 914-928, 2017.

[6] H. Wang and D. Suter, "A consensus-based method for tracking: Modelling background scenario and foreground appearance," *Pattern Recognit.*, vol. 40, pp. 1091-1105, 2007.

[7] O. Barnich and M. V. Droogenbroeck, "ViBe: A universal background subtraction algorithm for video sequences," *IEEE Trans. Image Process.*, vol. 20, no. 6, pp. 1709-1724, 2011.

[8] Z. Zivkovic, "Improved adaptive Gaussian mixture model for background subtraction," in *Proc. IEEE Int. Conf. Pattern Recognit.*, vol. 2, pp. 28-31, 2004.

[9] S. Varadarajan, P. Miller and H. Zhou, "Spatial mixture of Gaussians for dynamic background modelling," in *Proc. IEEE Int. Conf. Adv. Video Signal Based-Surveill.*, pp. 63-68, 2013.

[10] M. Mandal, P. Saxena, S. K. Vipparthi and S. Murala, "CANDID: Robust Change Dynamics and Deterministic Update Policy for Dynamic Background Subtraction," in *Proc. IEEE Int. Conf. Pattern Recognit.*, pp. 2468-2473, 2018.

[11] S. Jiang and X. Lu, "WeSamBE: A weight-sample-based method for background subtraction," *IEEE Trans. Circuits Systems for Video Technol.*, vol. 28, no. 9, pp. 2105-2115, 2018.

[12] M. Mandal, M. Chaudhary, S. K. Vipparthi, S. Murala, A. B. Gonde and S. K. Nagar, "ANTIC: ANTithetic Isomeric Cluster Patterns for Medical Image Retrieval and Change Detection," *IET Comput. Vis.*, vol. 13, no. 1, pp. 31-43, 2019.

[13] S. Hasan and S. C. S. Cheung, "Universal multimode background subtraction," *IEEE Trans. Image Process.*, vol. 26, no. 7, pp. 3249-3260, 2017.

[14] P. L. St-Charles, G. A. Bilodeau and R. Bergevin, "A Self-Adjusting Approach to Change Detection Based on Background Word Consensus," in *Proc. IEEE Winter Conf. Appl. Comput. Vis.*, pp. 990-997, 2015.

[15] M. Braham, S. Piérard and M. V. Droogenbroeck, "Semantic background subtraction," in *Proc. IEEE Int. Conf. Image Process.*, pp. 4552-4556, 2017.

[16] C. Lin, B. Yan and W. Tan, "Foreground Detection in Surveillance Video with Fully Convolutional Semantic Network," in *Proc. IEEE Int. Conf. Image Process.*, p. 4118-4122, 2018.

[17] L. A. Lim and H. Y. Keles, "Foreground Segmentation Using a Triplet Convolutional Neural Network for Multiscale Feature Encoding," arXiv preprint arXiv:1801.02225, 2018.

[18] D. Zeng and M. Zhu, "Multiscale Fully Convolutional Network for Foreground Object Detection in Infrared Videos," *IEEE Geosci. Remote Sens. Lett.*, vol. 15, no. 4, pp. 617-621, 2018.

[19] K. Lim, W. D. Jang and C. S. Kim, "Background subtraction using encoder-decoder structured convolutional neural network," in *Proc. IEEE Int. Conf. Adv. Video Signal Based-Surveill.*, pp. 1-6, 2017.

[20] T. P. Nguyen, C. C. Pham, S. V. U. Ha and J. W. Jeon, "Change Detection by Training a Triplet Network for Motion Feature Extraction," *IEEE Trans. Circuits Syst. Video Technol.*, vol. 29, no. 2, pp. 433-446, 2019.

[21] M. Babaee, D. T. Dinh and G. Rigoll, "A deep convolutional neural network for video sequence background subtraction," *Pattern Recognit.*, vol. 76, pp. 635-649, 2018.

[22] M. Braham and M. Van Droogenbroeck, "Deep background subtraction with scene-specific convolutional neural networks," in *Proc. IEEE Int. Conf. Syst., Signals and Image Process.*, pp. 1-4, 2016.

[23] Y. Gao, H. Cai, X. Zhang, L. Lan and Z. Luo, "Background subtraction via 3D convolutional neural networks," In *Proc. IEEE Int. Conf. Pattern Recognit.*, pp. 1271-1276, 2018.

[24] Y. Chen, J. Wang, B. Zhu, M. Tang and H. Lu, "Pixel-wise deep sequence learning for moving object detection," *IEEE Trans. Circuits Syst. Video Technol.*, to be published, doi: 10.1109/TCSVT.2017.277031.

[25] P. W. Patil and S. Murala, "MSFgNet: A Novel Compact End-to-End Deep Network for Moving Object Detection," *IEEE Trans. Intell. Transportation Systems*, to be published, doi: 10.1109/TITS.2018.2880096.

[26] L. Yang, J. Li, Y. Luo, Y. Zhao, H. Cheng and J. Li, "Deep Background Modeling Using Fully Convolutional Network," *IEEE Trans. Intell. Transportation Systems*, vol. 19, no. 1, pp. 254-262, 2018.

[27] M. C. Bakkay, H. A. Rashwan, H. Salmane, L. Khoudour, D. Puigtt and Y. Ruichek, "BSCGAN: Deep Background Subtraction with Conditional Generative Adversarial Networks," in *Proc. IEEE Int. Conf. Image Process.*, pp. 4018-4022, 2018.

[28] H. M. Heys and S. E. Tavares, "Avalanche characteristics of substitution-permutation encryption networks," *IEEE Trans. Computers*, vol. 44, no. 9, pp.1131-1139, 1995.

[29] Y. Wang, P. M. Jodoin, F. Porikli, J. Konrad, Y. Benezeth, and P. Ishwar, "CDnet 2014: an expanded change detection benchmark dataset," in *Proc. IEEE Conf. Comput. Vis. Pattern Recognit. Workshops*, pp. 387-394, 2014.